\documentclass[10pt,twocolumn]{article}

\usepackage[T1]{fontenc}
\usepackage[utf8]{inputenc}
\usepackage{newtxtext,newtxmath}
\usepackage[margin=0.72in,columnsep=0.25in]{geometry}
\usepackage{microtype}
\usepackage{amsmath}
\usepackage{booktabs}
\usepackage{array}
\usepackage{tabularx}
\usepackage{multirow}
\usepackage{graphicx}
\usepackage{xcolor}
\usepackage{enumitem}
\usepackage{caption}
\usepackage{subcaption}
\usepackage{natbib}
\usepackage{xurl}
\usepackage{hyperref}
\usepackage{tikz}
\usetikzlibrary{arrows.meta,positioning,shapes.geometric,fit,calc}

\definecolor{graphblue}{HTML}{24557A}
\definecolor{entitygreen}{HTML}{4A7C59}
\definecolor{memorygray}{HTML}{5D6670}
\definecolor{lightblue}{HTML}{EAF2F8}
\definecolor{lightgreen}{HTML}{EDF5EF}
\definecolor{lightgray}{HTML}{F1F3F5}
\definecolor{accentorange}{HTML}{B85C18}
\definecolor{lightorange}{HTML}{FFF1E6}

\hypersetup{
  colorlinks=true,
  linkcolor=graphblue,
  citecolor=entitygreen,
  urlcolor=graphblue,
  pdfauthor={Shumao Sun},
  pdftitle={Entity-Memory Graph Retrieval Improves Evidence Coverage in Long-Conversation Question Answering}
}

\setlist{nosep,leftmargin=*}
\newcommand{\recallacc}{\texttt{recall\_acc}}
\newcommand{\bembed}{B_{\mathrm{embed}}}
\newcommand{\topk}{\textit{top-k}}

\title{\textbf{Entity--Memory Graph Retrieval Improves Evidence Coverage\\
in Long-Conversation Question Answering}}
\author{Shumao Sun\\
\small Tsinghua University}
\date{}

\begin{document}
\maketitle

\begin{abstract}
Entity--Memory graph retrieval keeps dialogue turns as verbatim Memory nodes,
links repeated mentions through shared Entities, and connects adjacent
Memories with directed chronological edges. At query time the retriever moves
from Entity gating through semantic fusion and one-hop chronological recovery
to dense backfill. The path can keep a neighboring Memory that dense cosine ranking
would otherwise omit. A matched dense control
shares the Memory and query vectors, context budget, requested answer
protocol, and evaluator, isolating graph structure from changes to the
reader.

On 1,986 questions from ten LoCoMo conversations, graph retrieval raises
official evidence recall at \topk{} 25 from 79.7468\% to 84.4842\%. The recall
advantage is supported from \topk{} 5 to 50, while no matched cutoff supports
an overall final-answer F1 difference. Four paper-eligible requested
configurations support empirical robustness across the tested GPT-3.5 and
DeepSeek extractors on both outcomes. Embedding robustness is mixed: F1 has
no supported contrast, but recall is sensitive to the embedding artifact.
The comparison isolates a retrieval-coverage gain from graph structure. It
does not establish a final-answer F1 gain, model or embedding equivalence, or
cross-dataset generalization.
\end{abstract}

\noindent\textbf{Keywords:} long-term conversational memory; Entity--Memory
graph retrieval; evidence retrieval; LoCoMo

\section{Introduction}

Conversational agents accumulate facts, events, preferences, and relationships
over interactions that exceed a model's usable context. Long-term memory
systems must therefore decide what to store, how to organize it, and which
parts to retrieve for a new question. The LoCoMo benchmark makes this problem
concrete through long, multi-session conversations and questions requiring
single-hop, multi-hop, temporal, commonsense, and adversarial reasoning
\citep{maharana2024locomo}. Long-context and retrieval-augmented generation
improve memory-related question answering on LoCoMo, but long-range temporal
and causal reasoning remain difficult.

Dense retrieval provides a natural control. It encodes each conversational
Memory item, scores that item against the question, and sends the
highest-ranked items to an answer model. The design is simple and scalable,
but it treats Memory items as independent. A graph can represent repeated
entities and chronological adjacency, allowing retrieval to use structure
that is absent from a flat index. A valid comparison must distinguish that
structural contribution from changes in embeddings, context budgets, answer
models, or evaluation.

We evaluate an Entity--Memory graph built only from conversation content.
Each dialogue item becomes a Memory node, normalized Entities become shared
nodes, and chronological links connect neighboring Memories. Entity matching
is fused with semantic Memory similarity and chronological expansion at query
time. Test questions form retrieval queries only. Questions, answers, evidence
labels, categories, predictions, and evaluator output are excluded from graph
construction.

The study addresses three research questions:

\begin{enumerate}[label=\textbf{RQ\arabic*.}]
  \item When Memory nodes, ordered query vectors, answer generation, and
  evaluation are fixed, does complete graph retrieval improve evidence recall
  over matched dense retrieval, and does any retrieval gain translate into
  higher overall final-answer F1?
  \item Which retrieval components are associated with the observed
  difference in evidence recall?
  \item How robust is the graph-retrieval advantage across context cutoffs
  and input profiles, how sensitive are F1 and evidence recall to fusion and
  sequence settings and paper-eligible requested extractor/embedding
  configurations, and where are the category-level strengths and weaknesses
  concentrated?
\end{enumerate}

These questions motivate three contributions:

\begin{enumerate}[label=(\arabic*)]
  \item a formal conversation-only Entity--Memory retrieval model that
  separates graph construction from question-time retrieval;
  \item an all-question matched comparison that holds Memory vectors, ordered
  query vectors, answer generation, and evaluation fixed while measuring
  evidence recall separately from final-answer F1; and
  \item outcome-specific analyses of components, cutoffs, input profiles,
  parameters, the four paper-eligible requested configurations, descriptive
  category profiles, audited cases, and cold/warm operation that delimit where
  the retrieval difference holds and whether it reaches final-answer F1.
\end{enumerate}

Together, these contributions isolate the retrieval effect of graph structure
from changes to the reader or evaluator. The full experiment covers the
primary matched A/B comparison, structural ablations, four context cutoffs,
raw and time-annotated inputs, fusion/sequence sensitivity, and a
paper-eligible two-requested-extractor-by-two-embedding matrix. The main
pattern is outcome-specific: graph structure improves evidence coverage, the
tested requested extractors support empirical robustness in both F1 and
recall, and the tested embeddings yield mixed robustness, with empirical F1
robustness but supported recall differences.

\section{Related Work}

\subsection{Why long conversations need structured retrieval}

Long conversations scatter related facts across sessions, so a memory system
must preserve both semantic relevance and the relations among separated
turns. Retrieval-augmented generation combines a parametric generator with retrieved
non-parametric evidence \citep{lewis2020rag}. Dense Passage Retrieval showed
that a learned dual encoder can retrieve candidate passages for open-domain
question answering using dense representations
\citep{karpukhin2020dpr}. These studies establish the flat dense-retrieval
control used in our matched comparison.

Later systems organize long inputs above the individual-chunk level. RAPTOR
recursively embeds, clusters, and summarizes text into a retrieval tree
\citep{sarthi2024raptor}. MemoRAG uses a long-range model to form global memory
and generate clues for a downstream retriever \citep{qian2024memorag}. Our
method also introduces structure above independent Memory vectors, but it
retains dialogue items as the answer evidence rather than replacing them with
generated summaries.

\subsection{Long-term conversational memory}

LoCoMo introduced a human-verified benchmark of long, persona-grounded,
multimodal conversations and evaluates question answering, event
summarization, and dialogue generation \citep{maharana2024locomo}.
LongMemEval later framed assistant memory as indexing, retrieval, and reading,
and evaluates information extraction, multi-session reasoning, temporal
reasoning, knowledge updates, and abstention \citep{wu2025longmemeval}. These
benchmarks motivate pipeline-level evaluation rather than attributing all
downstream behavior to the language model.

Early agent-memory architectures store experience for later retrieval and
reflection. Generative Agents records observations, derives higher-level
reflections, and retrieves memories for planning
\citep{park2023generative}. MemoryBank adds persistent conversational memory
with retrieval, updating, and time-dependent forgetting
\citep{zhong2024memorybank}. MemGPT instead manages limited context through
virtual memory and hierarchical tiers \citep{packer2023memgpt}.

Recent systems place more structure in the memory store. Mem0 extracts,
consolidates, and retrieves salient conversational information, with a graph
variant for relational memory \citep{chhikara2025mem0}. A-Mem dynamically
indexes and links structured memory notes into an evolving network
\citep{xu2025amem}. Our study examines a narrower question than these
end-to-end memory architectures. Given a fixed requested reader protocol and
fixed Memory representations, we test whether a conversation-derived graph
changes evidence retrieval relative to a flat dense control.

\subsection{Graph retrieval and the comparison gap}

Graph retrieval can expose relations that independent text chunks omit.
HippoRAG combines a knowledge graph with Personalized PageRank for long-term
knowledge integration and multi-hop retrieval
\citep{gutierrez2024hipporag}. GraphRAG constructs a graph index and community
summaries for global questions over document collections
\citep{edge2024graphrag}. GRAG retrieves textual subgraphs and supplies textual
and topological views to a generator \citep{hu2025grag}, while LightRAG
combines graph indexing with dual-level retrieval over entities and relations
\citep{guo2025lightrag}.

Graph structure has also become central to conversational memory. APEX-MEM
stores temporally grounded events in a property graph and resolves evolving
information at retrieval time \citep{banerjee2026apexmem}. Mnemis combines a
base similarity graph with top-down selection over a semantic hierarchy
\citep{tang2026mnemis}. GAM separates an event-progression graph from a
topic-association network \citep{wu2026gam}, whereas TiMem consolidates
conversation observations into a temporal hierarchy
\citep{li2026timem}. Our graph is smaller in scope. Shared Entity nodes link
mentions across dialogue-item Memories, and chronological edges preserve local
sequence. These systems motivate graph memory, but they also expose an
attribution problem: system-level comparisons often change the graph,
encoder, reader, prompt, and context budget together. DRAGON, for example, is
a dense retriever trained with diverse data augmentation
\citep{lin2023dragon} and is used in the original LoCoMo RAG setup. We
therefore treat external systems as descriptive anchors and use an internal
matched dense control to isolate graph retrieval. This is an
experimental-control choice, not a new evaluation metric.

\section{Method}

\subsection{Task and estimands}

Let a conversation be an ordered collection of Memory items
\(\mathcal{M}=\{m_1,\ldots,m_n\}\), where each item contains a dialogue
identifier, speaker, text, timestamp, and optional image caption. Given a
question \(q\), a retriever returns an ordered context
\(R_k(q)\subseteq\mathcal{M}\) with \(|R_k(q)|=k\). The study tests whether a
conversation-derived graph changes the evidence contained in \(R_k(q)\) when
the Memory representations, query vectors, reader, and evaluator are fixed.

Graph construction is a function of conversation content only:
\begin{equation}
G=B(\mathcal{M})=(V_M\cup V_E,\ E_{EM}\cup E_{MM}).
\label{eq:graph}
\end{equation}
Here, \(V_M\) contains one node per Memory item, \(V_E\) contains normalized
Entities shared across items, \(E_{EM}\) contains Entity--Memory mention
links, and \(E_{MM}\) contains directed next/previous links between
chronologically adjacent Memories. The constructor \(B\) has no access to
questions, answers, evidence annotations, category labels, predictions, or
evaluator output. Questions enter only through the retrieval function
\(R_k(q;G)\).

The primary estimand is the paired difference in official evidence recall
between graph retrieval and matched dense retrieval. Final-answer F1 is a
separate downstream outcome. This distinction prevents additional retrieved
evidence from being interpreted automatically as an improvement in generated
answers.

\begin{figure*}[t]
\centering
\resizebox{\textwidth}{!}{%
\begin{tikzpicture}[
  x=1cm,y=1cm,
  box/.style={draw,rounded corners=2pt,minimum height=0.62cm,
    align=center,font=\scriptsize,inner xsep=4pt},
  memory/.style={box,fill=lightgray,draw=memorygray},
  entity/.style={circle,draw=entitygreen,fill=lightgreen,inner sep=1pt,
    font=\scriptsize\bfseries,minimum size=0.56cm},
  stage/.style={box,fill=lightorange,draw=accentorange,very thick,
    font=\tiny,minimum height=0.72cm},
  densebox/.style={box,fill=lightblue,draw=graphblue,font=\tiny},
  miss/.style={box,fill=white,draw=memorygray,dashed,font=\tiny},
  hit/.style={box,fill=lightorange,draw=accentorange,very thick,font=\tiny},
  reader/.style={box,fill=white,draw=graphblue,very thick,font=\tiny},
  proposed/.style={-{Latex[length=1.8mm]},very thick,draw=accentorange},
  control/.style={-{Latex[length=1.8mm]},thick,dashed,draw=memorygray},
  chrono/.style={-{Latex[length=1.6mm]},very thick,draw=entitygreen}
]
\fill[lightgray!45,rounded corners=3pt] (-0.20,-0.55) rectangle (6.55,4.35);
\node[anchor=west,font=\scriptsize\bfseries,text=memorygray] at (-0.05,4.08)
  {(a) Ordinary dense RAG};
\node[memory,minimum width=1.25cm] (am1) at (1.05,3.22) {$m_{t-1}$};
\node[memory,minimum width=1.25cm] (am2) at (3.20,3.22) {$m_{t}$};
\node[memory,minimum width=1.25cm] (am3) at (5.35,3.22) {$m_{t+1}$};
\node[draw=memorygray,dashed,rounded corners=3pt,inner sep=5pt,
  fit=(am1)(am2)(am3)] (aindex) {};
\node[font=\tiny,text=memorygray] at (3.20,2.48)
  {flat Memory index; no Entity or edge};
\node[densebox] (aq) at (1.05,1.48) {question\\vector};
\node[densebox] (acos) at (3.20,1.48) {cosine\\ranking};
\node[miss] (atop) at (5.35,1.48) {RAG top-\(k\)\\may drop $m_{t+1}$};
\draw[control] (aindex) -- (acos);
\draw[control] (aq) -- (acos);
\draw[control] (acos) -- (atop);
\node[reader] (areader) at (3.20,0.12) {same reader};
\draw[control] (atop) -- (areader);

\fill[lightorange!28,rounded corners=3pt] (6.75,-0.55) rectangle (17.55,4.35);
\node[anchor=west,font=\scriptsize\bfseries,text=accentorange] at (6.90,4.08)
  {(b) Entity--Memory graph retrieval};
\node[entity] (e1) at (9.40,3.52) {$e_1$};
\node[entity] (e2) at (11.60,3.52) {$e_2$};
\node[memory,minimum width=1.15cm] (bm1) at (8.40,2.58) {$m_{t-1}$};
\node[memory,minimum width=1.15cm] (bm2) at (10.50,2.58) {$m_{t}$};
\node[memory,minimum width=1.15cm] (bm3) at (12.60,2.58) {$m_{t+1}$};
\draw[chrono] (bm1) -- (bm2);
\draw[chrono] (bm2) -- (bm3);
\draw[entitygreen,thick] (e1) -- (bm1);
\draw[entitygreen,thick] (e1) -- (bm2);
\draw[entitygreen,thick] (e2) -- (bm2);
\draw[entitygreen,thick] (e2) -- (bm3);
\node[font=\tiny,text=entitygreen,align=left] at (14.85,3.10)
  {shared Entities +\\chronological edges};
\node[densebox] (bq) at (8.10,1.15) {question\\Entities + vector};
\node[stage] (gate) at (10.05,1.15) {\textbf{1} Entity\\gate};
\node[stage] (fusion) at (11.75,1.15) {\textbf{2} semantic\\fusion};
\node[stage] (seq) at (13.45,1.15) {\textbf{3} one-hop\\recovery};
\node[stage] (back) at (15.15,1.15) {\textbf{4} dense\\backfill};
\node[hit] (gctx) at (16.75,1.15) {graph top-\(k\)\\keeps $m_{t+1}$};
\draw[proposed] (bq) -- (gate);
\draw[proposed] (gate) -- (fusion);
\draw[proposed] (fusion) -- (seq);
\draw[proposed] (seq) -- (back);
\draw[proposed] (back) -- (gctx);
\node[reader] (breader) at (16.75,0.12) {same reader};
\draw[proposed] (gctx) -- (breader);
\node[font=\tiny,text=accentorange] at (13.45,0.38)
  {recovers the adjacent fact};
\end{tikzpicture}%
}
\caption{Matched contrast between ordinary dense RAG and Entity--Memory
graph retrieval.
(a)~The control ranks independent Memory vectors by cosine similarity and
sends the top-\(k\) items to the reader; it has no Entity nodes and no
chronological edges.
(b)~Conversation turns first become a graph with shared Entities and
directed Memory edges; retrieval then gates, fuses, recovers one-hop
neighbors, and backfills. Both paths share Memory vectors, the query
vector, \topk{}, the requested answer protocol, and the frozen evaluator.
The neighbor annotation is schematic and does not report a metric. The
control is the internal matched dense condition, not the official LoCoMo
DRAGON setup.}
\label{fig:pipeline}
\end{figure*}
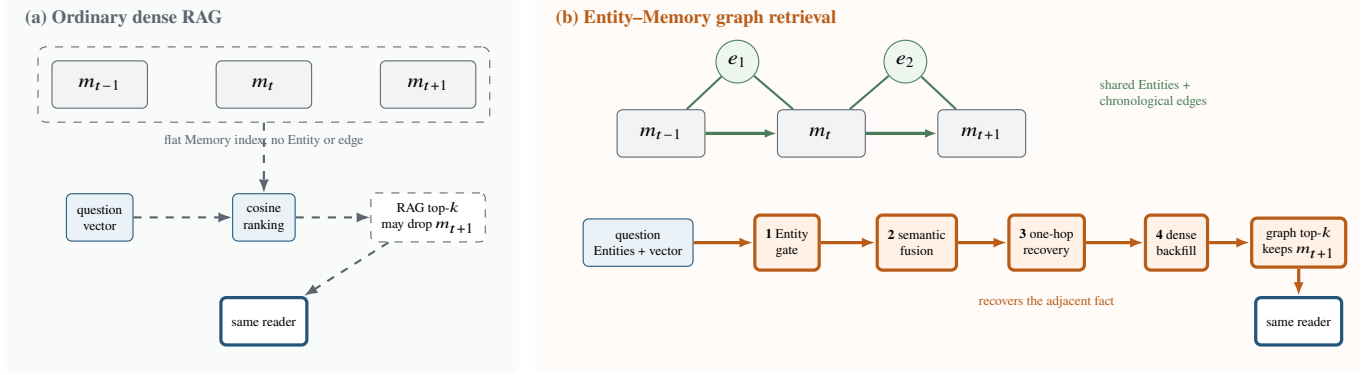

\subsection{Conversation-only graph construction}

Figure~\ref{fig:pipeline}(a) is the matched dense RAG control: independent
Memory ranking with no Entity or chronological signal.
Figure~\ref{fig:pipeline}(b) shows the conversation-only graph and the
progressive retrieval path. For each
conversation, the constructor consumes session timestamps, dialogue
identifiers, speaker names, dialogue text, and image captions. Each normalized
dialogue item becomes a Memory node. A \texttt{gpt-3.5-turbo} extractor at
temperature 0.3 identifies Entities from the conversation text. Entity strings
are normalized, deduplicated case-insensitively, and connected to every Memory
in which they occur. Deterministic speaker links add the dialogue speaker as an
Entity. Adjacent Memory nodes are connected in chronological order.

The complete ten-conversation graph contains 5,882 Memory nodes. Graph
construction does not consume QA questions, gold answers, evidence
annotations, category labels, judge outputs, previous predictions, or
question-driven ledgers. This separation prevents test questions from shaping
the stored memory structure.

\subsection{Entity-gated retrieval with chronological recovery}

The question supplies two retrieval-only signals. A question-Entity extractor
produces a set of normalized keys \(Q_E(q)\), and the question vector is loaded
from a complete artifact generated with OpenAI's
\texttt{text-embedding-3-small} model~\citep{openaiTextEmbedding3Small}. The
stored artifact is immutable and bound to the dataset hash, ordered question
digests, model, role, vector dimension, normalization, and artifact SHA-256.
Formal retrieval fails on a cache miss and makes no live query-embedding
request.

Let \(\rho(e,u)\in[0,1]\) be the peak-normalized BM25 score between Entity
node \(e\) and question key \(u\). Exact normalized-key equality receives a
score of one. Matches below threshold \(\tau=0.5\) are removed, and at most
20 Entity nodes are retained for each key. For the set \(Q_E^+(q)\) of keys
with at least one retained match, the degree-discounted Entity score is
\begin{equation}
\begin{aligned}
h(e,q)={}&\frac{1}{|Q_E^+(q)|}
\sum_{u\in Q_E^+(q)}\rho(e,u)\\
&{}\times\frac{1}{\log(1+\deg(e))}.
\end{aligned}
\label{eq:entity}
\end{equation}
For Memory \(m\), let \(c(m,q)\) and \(w(m,q)\) be the maximum incident
scores from content Entities and person/speaker Entities, respectively. The
Memory seed score uses \(c_m=c(m,q)\), \(w_m=w(m,q)\), and \(\delta=0.25\):
\begin{equation}
s_e^{(0)}(m,q)=
\begin{cases}
\max\{c_m,w_m\}, & c_m>0,\\
\delta w_m, & c_m=0.
\end{cases}
\label{eq:seed}
\end{equation}
Thus, a speaker match is dampened only when no content Entity supports the
same Memory.

One-hop chronological expansion transfers each seed score to its immediately
preceding and following Memories. With \(N_{\mathrm{seq}}(m)\) denoting those
neighbors, the expanded Entity score is
\begin{equation}
\begin{aligned}
s_e(m,q)=\max\biggl\{&s_e^{(0)}(m,q),\\
&\lambda\max_{m'\in N_{\mathrm{seq}}(m)}s_e^{(0)}(m',q)\biggr\}.
\end{aligned}
\label{eq:sequence}
\end{equation}
In the experiments, we generally set \(\lambda\) to 0.5; the preregistered
sensitivity analysis also tests 0.25 and 1.0.

The nonempty Entity scores define the gated candidate set
\(\mathcal{C}(q)=\{m:s_e(m,q)>0\}\). If this set is empty, the candidate set
is all Memories. Within the active set, the primary condition combines the
Entity score with signed cosine similarity \(s_s(m,q)\):
\begin{equation}
s(m,q)=\alpha s_e(m,q)+\beta s_s(m,q).
\label{eq:fusion}
\end{equation}
The primary setting uses \(\alpha=0.30\) and \(\beta=0.70\).
Candidates are ordered by decreasing \(s(m,q)\) and then by dialogue id. When
the gated set contains fewer than \(k\) Memories, unused items from full-pool
dense retrieval fill the remaining positions without displacing gated items.

The primary context budget (\topk{}) is 25. Cutoff analyses use \topk{} 5, 10,
and 50. The dense control ranks the same Memory representations with the same
semantic query vectors but does not use Entity matching or sequence expansion.
A dense graph-construction control, \(\bembed\), verifies that graph
construction alone does not change dense retrieval.

\section{Experimental Setup}

\subsection{Matched conditions and comparison boundary}

The primary comparison holds constant the dataset and row order, Memory text,
Memory vectors, immutable ordered query vectors, answer prompt, answer model,
token budget, evaluation package, output isolation, and aggregation. The
treatment activates Entity matching, graph fusion, and chronological
expansion. The \(\bembed\) control must reproduce the dense control's ordered
context identifiers for all rows before the primary graph contrast is
interpreted.

The original LoCoMo RAG setup uses DRAGON, but our pinned upstream
reproduction did not pass the declared tolerance. The local raw-DRAGON
condition is therefore diagnostic and is excluded from the matched and
inferential comparisons. The primary evidence concerns A, \(\bembed\), and B
under the shared artifacts defined above.

Each metric-bearing condition writes to an absent, condition-specific output
directory. Its record includes the source commit, resolved configuration,
model names, artifact identities, query-cache hits and misses, prompt budgets,
and graph compliance. The vendored evaluator is verified against its SHA-256
manifest before formal use.

\subsection{Ablation study design}

The design matrix separates structural attribution, operating-condition
stress tests, local parameter sensitivity, and model substitution.
Within each family, the listed treatment changes while the comparison boundary
retains the shared artifacts and evaluation protocol; the final row records a
completed condition that is useful diagnostically but not eligible for
model-only inference.

\begin{table*}[!t]
\centering
\caption*{Ablation and robustness design. Each row changes only the listed
factor within its declared comparison; all conditions retain the same
requested answer protocol and frozen evaluator.}
\begin{tabularx}{\textwidth}{@{}p{0.17\textwidth}p{0.31\textwidth}Xp{0.23\textwidth}@{}}
\toprule
Family & Treatment and control & Fixed factors & Purpose \\
\midrule
Structural components & Remove sequence expansion, Entity-score fusion,
semantic scoring, or speaker links from complete B & Dataset, Memory text,
primary vectors, \topk{} 25, requested reader protocol, frozen evaluator & Attribute the primary
retrieval difference \\
Cutoff and input & Matched A/B at \topk{} 5, 10, 25, 50; raw-text versus
time-annotated input & Paired rows and within-profile artifacts & Test context
budget and input-profile dependence \\
Parameters & Entity/Semantic weights 0.10/0.90, 0.30/0.70, 0.50/0.50;
sequence scale 0.25, 0.5, 1.0 & Complete-B graph and reader & Bound local
sensitivity without claiming universal optima \\
Robustness across configurations & Requested \texttt{gpt-3.5-turbo} or
\texttt{deepseek-v4-flash} Entity extraction crossed with TES or Doubao
embeddings & Complete B,
\topk{} 25, requested extraction budget 2,500, requested GPT-3.5 reader protocol & Test extractor and
embedding substitutions in a paper-eligible $2\times2$ matrix with
outcome-specific robustness claims \\
Diagnostic only & Two \texttt{gpt-5-mini} extraction cells & Reader and
retrieval settings fixed, but runtime extraction budget was 4,000 rather than
the frozen 2,500 & Excluded from model-only inference pending rerun \\
\bottomrule
\end{tabularx}
\end{table*}

\subsection{Evaluation protocol and statistical design}

Retrieved Memory text and identifiers enter the frozen LoCoMo-aligned answer
interface. The requested answer model is \texttt{gpt-3.5-turbo}, with a
system-role prompt, temperature 0, one question per batch, and a 32-token
completion limit. The immutable evaluation package retains its
category-specific generation, decoding, F1, evidence-recall, rounding, and
aggregation behavior. Overall F1 measures final-answer overlap, whereas
\recallacc{} is the official evidence-recall definition. The
repository-defined Categories 1--4 subset F1 is reported only as a diagnostic,
not as an official LoCoMo metric.

Every formal condition contains the same 1,986 QA rows from ten conversations.
Matched differences are estimated with 10,000 paired-question bootstrap
resamples and 10,000 resamples of whole conversations, using seed 20260727.
The conversation-cluster estimator reflects uncertainty over the ten
conversation units. Component, input, fusion, and sequence families use Holm
step-down correction within each outcome and estimator. A difference is
treated as supported only when the preregistered evidence gate passes. Failure
to reject is not interpreted as equivalence.

The paper-eligible configuration matrix crosses two requested Entity
extractors with two embedding aliases and uses 10,000 whole-conversation
bootstrap resamples with seed 20260814. Holm correction is applied separately
to extraction/F1, extraction/recall, embedding/F1, and embedding/recall. The
preregistered sequential stop gate uses an absolute three-point band and is
triggered only when a 95\% cluster interval lies wholly below $-3$ points or
wholly above $+3$ points. We report empirical robustness only as an
outcome-specific interpretation when a tested substitution has no
Holm-corrected rejection and does not trigger this gate; this is not a formal
equivalence test. Two completed
\texttt{gpt-5-mini} cells are retained only as diagnostic observations: their
frozen records specify a 2,500-token extraction limit, whereas the executed
model-specific path used 4,000 tokens and 12 calls exceeded 2,500. They are
therefore excluded from the model-only robustness claims and category ranges.
For all F1 contrasts, the requested GPT-3.5 alias, prompt, role, temperature,
and token budget are fixed, but the provider-returned deployment revision was
not recorded and Category-5 option order remains unseeded. We therefore treat
F1 as an observed downstream outcome under a matched requested protocol;
\recallacc{} provides the cleaner retrieval-stage comparison.

\section{Results}

\subsection{Primary matched comparison}

\begin{table}[t]
\centering
\caption{Primary matched comparison at \topk{} 25. Values are percentages.}
\label{tab:primary}
\begin{tabular}{@{}lrr@{}}
\toprule
Condition & Overall F1 & \recallacc{} \\
\midrule
A (dense Memory) & 42.0681 & 79.7468 \\
\(\bembed\) (dense graph control) & 42.0677 & 79.7468 \\
B (complete graph retrieval) & \textbf{42.5680} & \textbf{84.4842} \\
\bottomrule
\end{tabular}
\end{table}

Complete graph retrieval improves \recallacc{} over the matched dense control
by 4.7374 percentage points (Table~\ref{tab:primary}). The paired-question
95\% interval is 3.6504--5.8425 points, and the conversation-cluster interval
is 3.5286--6.0124 points. Both two-sided \(p\)-values are 0.0002. Overall F1
changes by 0.4998 points, with paired and cluster intervals of
\(-0.5745\)--1.5514 and \(-0.1773\)--1.2711 points. The F1 difference is not
supported by either estimator.

The dense graph-construction control and A have zero ordered-context
mismatches across all 1,986 rows and identical evidence recall. Their F1
values differ by less than 0.001 percentage points, consistent with frozen
Category-5 option-order randomness rather than a retrieval difference.

\subsection{Component analysis}

\begin{table*}[t]
\centering
\caption{Component and input contrasts at \topk{} 25. Differences are
treatment $-$ control in percentage points. ``Supported'' refers to the
preregistered family-corrected inference unless noted otherwise.}
\label{tab:components}
\begin{tabularx}{\textwidth}{@{}Xrrl@{}}
\toprule
Contrast & \(\Delta\)F1 & \(\Delta\)\recallacc{} & Conclusion \\
\midrule
B $-$ B without sequence expansion & \(+0.5137\) & \(+1.6728\) & Recall supported; F1 not supported \\
B $-$ B without Entity-score fusion & \(+0.8014\) & \(+4.7626\) & Recall supported; F1 not supported \\
Sequence expansion after Entity gating & \(-0.0609\) & \(+0.4364\) & Recall supported; F1 not supported \\
B $-$ Entity-only retrieval & \(+5.0780\) & \(+17.4634\) & Both outcomes supported \\
Raw-text B $-$ annotated-input B & \(+0.2407\) & \(-0.2472\) & Neither outcome supported \\
Raw-text B $-$ raw-text A & \(+0.3508\) & \(+4.9522\) & Recall supported; F1 not supported \\
B $-$ B without speaker links & \(+0.2415\) & \(+1.4822\) & Recall supported; F1 not supported \\
\bottomrule
\end{tabularx}
\end{table*}

The component analysis associates several graph operations with the retrieval gain
(Table~\ref{tab:components}). Complete B improves recall over the no-sequence
condition by 1.6728 points and over the no-Entity-score-fusion condition by
4.7626 points after Holm correction. Neither contrast supports F1. Sequence
expansion after Entity gating contributes 0.4364 recall points. The semantic
channel relative to Entity-only retrieval contributes 17.4634 recall points
and 5.0780 F1 points, with both outcomes supported within the component
family. Removing deterministic speaker links lowers recall by 1.4822 points
without a supported F1 change.

The apparently opposite signs for ``sequence expansion after Entity gating''
reflect different outcomes, not a contradiction. For the Category-4 question
``How long have Mel and her husband been married?'', the gated control misses
gold turn D3:16 and produces an incorrect verbose sentence that nevertheless
shares enough tokens with the reference to receive F1 0.429. Sequence
expansion retrieves D3:16 and yields the correct concise answer ``5 years'',
raising evidence recall from 0 to 1 but receiving F1 0.364. This case
illustrates why the aggregate $+0.4364$ recall-point contrast can coexist with
an unsupported $-0.0609$ F1-point contrast: token overlap and evidence
coverage measure different stages of the pipeline.

\subsection{Cutoff and input-profile robustness}

\begin{table*}[t]
\centering
\caption{Matched cutoff analysis. Values are percentages; differences are
B minus A in percentage points. Every paired-question and
conversation-cluster recall interval is above zero. Every corresponding F1
interval includes zero.}
\label{tab:cutoffs}
\begin{tabular}{@{}rrrrrrrll@{}}
\toprule
\topk{} & A F1 & B F1 & \(\Delta\)F1 & A recall & B recall &
\(\Delta\)recall & Paired recall CI & Cluster recall CI \\
\midrule
5  & 39.9053 & 39.8264 & \(-0.0790\) & 59.3584 & 64.1667 & \(+4.8083\) & [3.3450, 6.2808] & [2.8809, 6.6134] \\
10 & 41.7607 & 42.1412 & \(+0.3805\) & 68.9145 & 74.4847 & \(+5.5702\) & [4.2928, 6.8624] & [4.2365, 6.7522] \\
25 & 42.0681 & 42.5680 & \(+0.4998\) & 79.7468 & 84.4842 & \(+4.7374\) & [3.6504, 5.8425] & [3.5286, 6.0124] \\
50 & 42.0782 & 41.7832 & \(-0.2950\) & 86.7148 & 90.3306 & \(+3.6159\) & [2.7754, 4.4735] & [2.4603, 4.6426] \\
\bottomrule
\end{tabular}
\end{table*}

The evidence-recall advantage persists across all preregistered cutoffs
(Table~\ref{tab:cutoffs}). B minus A recall differences range from 3.6159 to
5.5702 points. Paired-question and conversation-cluster intervals are above
zero at every cutoff. Both matched F1 intervals include zero at every cutoff.
The cutoff analysis therefore supports an evidence-coverage advantage over the
tested range, not an answer-quality advantage.

The input-profile contrasts preserve the distinction between evidence recall
and final-answer F1 (Table~\ref{tab:components}). Time annotation has no
supported within-B overall effect after correction. When both A and B use raw
text, B retains a supported 4.9522-point recall advantage without an F1
advantage.

\subsection{Parameter sensitivity}

The parameter results bound the component findings in
Table~\ref{tab:components}. Reducing Entity/Semantic weights from 0.30/0.70 to
0.10/0.90 lowers recall by 2.3062 points after correction. Equal 0.50/0.50
weights are not distinguishable from the primary setting. No overall F1 or
recall contrast among sequence scales 0.25, 0.5, and 1.0 survives family
correction.

\subsection{Outcome-specific robustness across requested configurations and descriptive category profile}

\begin{table}[t]
\centering
\caption{Paper-eligible $2\times2$ configuration audit for complete B at
\topk{} 25. TES is
\texttt{text-embedding-3-small}; DB is
\texttt{doubao-embedding-vision}. Values are percentages.}
\label{tab:model-robustness}
\small
\begin{tabularx}{\columnwidth}{@{}Xlrr@{}}
\toprule
Requested extractor & Embedding alias & F1 & \recallacc{} \\
\midrule
GPT-3.5 & TES & 42.5680 & 84.4842 \\
DeepSeek v4 Flash & TES & 42.1087 & 84.6005 \\
GPT-3.5 & DB & 42.1507 & 82.1978 \\
DeepSeek v4 Flash & DB & 42.8158 & 82.5778 \\
\bottomrule
\end{tabularx}
\end{table}

Across the four protocol-matched cells in
Table~\ref{tab:model-robustness}, F1 spans 42.1087--42.8158. Neither the two
requested-extractor F1 contrasts nor their recall contrasts reject zero after
within-family Holm correction. These results support limited empirical
robustness of both outcomes across the tested GPT-3.5 and DeepSeek requested
extraction configurations, without establishing strict extractor
equivalence. The corresponding embedding F1 contrasts also do not reject
zero, but the DB frozen artifacts lower recall by 2.0227 points under
DeepSeek extraction and
2.2864 points under GPT-3.5 extraction. Both recall contrasts remain supported
after correction. Each 95\% cluster interval contains values on both sides of
the preregistered $-3$-point boundary, so neither interval triggers the gate
for material robustness failure. Passing that gate does not establish
embedding equivalence. Robustness across the tested embeddings is therefore
outcome-specific: F1 supports empirical robustness under the declared rule,
whereas evidence recall is sensitive to the frozen embedding artifacts.

Row-level examples show why the aggregate conclusion must remain
outcome-specific. With TES fixed, replacing GPT-3.5 extraction by DeepSeek on
``What gifts has Deborah received?'' recovers one of five annotated turns
(recall 0 to 0.2), yet the independently generated answer changes from ``A
bouquet from a friend'' to ``A bouquet'' and token F1 falls from 0.44 to 0,
partly because the gold answer misspells \emph{bouquet}. Conversely, with the
DeepSeek graph fixed, TES retrieves both annotated turns for the Voyageurs
National Park question (recall 1), whereas DB drops one (recall 0.5). DB's
incorrect ``Unnamed national park'' nevertheless receives F1 0.667, above the
TES answer's 0.138, because it shares a larger fraction of tokens with the
gold name. These cases do not establish single-row causal effects; they
document how stable aggregate F1 can coexist with observable retrieval
changes.

\begin{table*}[t]
\centering
\caption{Category profile across the four paper-eligible configurations, with descriptive
external anchors from the original LoCoMo Dialog@25 condition
\citep{maharana2024locomo} and LightGMEM
\citep{anonymous2026lightgmem}. External readers, embeddings, and category
scope differ, so the table does not establish cross-paper superiority.}
\label{tab:categories}
\begin{tabularx}{\textwidth}{@{}lrrrrrX@{}}
\toprule
Category & \multicolumn{1}{c}{B F1 range} & \multicolumn{1}{c}{Dialog F1} &
\multicolumn{1}{c}{LightGMEM F1} & \multicolumn{1}{c}{B recall range} &
\multicolumn{1}{c}{Dialog recall} & Descriptive position \\
\midrule
Multi-hop & 37.6--39.9 & 38.7 & 41.9 & 61.5--67.3 & 62.5 & B F1 range overlaps the Dialog anchor \\
Temporal & 40.2--43.1 & 37.2 & 56.4 & 89.2--90.2 & 83.5 & B recall is numerically above the Dialog anchor \\
Open-domain & 17.3--18.6 & 25.0 & 27.2 & 48.4--56.4 & 52.6 & Lowest B answer F1 range among Categories 1--4 \\
Single-hop & 63.8--64.1 & 59.9 & 64.1 & 91.1--92.7 & 87.1 & Highest B answer F1 range \\
Adversarial & 8.3--11.2 & 12.8 & --- & 80.2--82.6 & 66.3 & B recall is numerically above Dialog; raw B F1 is lowest under the separate Category-5 branch \\
\bottomrule
\end{tabularx}
\end{table*}

Concrete rows illustrate this profile. On the temporal question ``Which classes
did Evan join in mid-August 2023?'', dense A misses D8:12 and answers
``winter activities'' (recall 0, F1 0.2). B retrieves D8:12 together with its
D8:13--14 neighbors and answers ``Painting classes'' (recall 1, F1 1.0).
For the single-hop question asking what keeps Evan busy while his knee heals,
A retrieves nearby D11 turns but omits D11:6 and answers ``Swimming''; B adds
D11:6 at rank 19 and answers ``Watercolor painting'', moving both outcomes
from 0 to 1. These cases are consistent with Entity links locating the
relevant session and chronological recovery adding the exact adjacent fact;
they do not isolate either operation as the cause of a single-row outcome.

The weaker categories exhibit different retrieval and answer patterns. For the open-domain
Dr.~Seuss question, both A and B retrieve the annotated evidence that Caroline
collects classic children's books, but their answers omit the reference
rationale and receive F1 0.222 and 0.118. In these stored outputs, retrieving
the annotated evidence is insufficient to match the reference rationale.
Conversely, for the adversarial question about
what inspired Melanie's art-show painting, B retrieves D9:16 whereas A does
not, yet both stored outputs receive F1 0 under the frozen Category-5 branch.
This row is consistent with the aggregate pattern of higher adversarial
recall without an answer-quality advantage; it does not identify the
generation or scoring branch as the cause.

\subsection{Operational cost under cold and warm retrieval}

\begin{table*}[t]
\centering
\caption{Primary-B cold and warm measurement, separated into provider work
and retrieval latency. Cold stage times overlap and must not be summed as
end-to-end latency.}
\label{tab:cost}
\begin{tabular}{@{}lrrrrr@{}}
\toprule
Provider stage & Cold req. & Input tokens & Output tokens & Cold call time & Warm req. \\
\midrule
Conversation Entity extraction & 5,873 & 3,731,393 & 425,637 & 9,792.76 s & 0 \\
Memory embedding & 591 & 214,229 & 0 & 443.38 s & 0 \\
Question Entity extraction & 1,974 & 1,214,259 & 104,272 & 2,970.83 s & 0 \\
\bottomrule
\end{tabular}

\vspace{0.45em}
\begin{tabular}{@{}lrrr@{}}
\toprule
Retrieval state (1,986 QA) & Provider req. & Mean per QA & p95 per QA \\
\midrule
Cold & 0 & 1.5123 s & 2.6132 s \\
Warm & 0 & 0.004886 s & 0.006270 s \\
\bottomrule
\end{tabular}
\end{table*}

The warm replay makes no new provider requests in the three cache-sensitive
stages (Table~\ref{tab:cost}). The cold-to-warm retrieval speedup is 309.48$\times$.
Cold and warm each cover the same 44 ordered batches, with 1,998 reads from the
immutable query artifact, zero misses, and zero live query embeddings. This
includes a 1.29-second warm Memory-index cache load. The zero-request result
applies to replaying the same question set with complete
graph, index, query-vector, and question-Entity caches; unseen questions may
require new query embedding and question-Entity calls. The formal answer trace
contains 1,986 requests, 2,744,099 input tokens, and 16,269 output tokens.
Because answer generation is independent of cache warmth, the same validated
trace is attached to each state rather than rerun.

Graphs occupy 21,689,931 bytes, Memory indexes 29,946,660 bytes, and
Entity/question caches 3,411,176 bytes. Including the attached answer trace,
the measured cold path uses 7,903,980 input and 546,178 output tokens, or
8,450,158 total tokens. Amortized over 1,986 questions, this is 4,254.9 tokens
per question; the answer stage accounts for 1,389.9 tokens per question.
Applying OpenAI public list prices accessed on 1 August 2026 gives an
illustrative total of \$4.67, or \$0.00235 per question.\footnote{Prices used
for this contextual conversion were \$0.50/\$1.50 per million GPT-3.5 Turbo
input/output tokens and \$0.02 per million text-embedding-3-small tokens:
\url{https://developers.openai.com/api/docs/models/gpt-3.5-turbo} and
\url{https://developers.openai.com/api/docs/models/text-embedding-3-small}.}
This conversion excludes the separately built read-only query-vector artifact,
storage and platform fees, and unrecorded external charges. It is not a frozen
experimental metric.

\section{Discussion}

The experiments separate evidence coverage from answer quality. Entity links,
semantic scoring, and chronological expansion retrieve more annotated
evidence than the matched dense control, and the difference persists across
four context budgets. Under the matched requested answer protocol, the
corresponding overall F1 difference is not supported. Evaluations that report
only answer scores can therefore obscure retrieval changes, while retrieval
improvements should not be described as end-to-end quality gains.

This separation complements LongMemEval's decomposition of assistant memory
into indexing, retrieval, and reading stages \citep{wu2025longmemeval}.
Systems such as Mem0, A-Mem, APEX-MEM, Mnemis, GAM, and TiMem jointly change
several of these stages
\citep{chhikara2025mem0,xu2025amem,banerjee2026apexmem,tang2026mnemis,wu2026gam,li2026timem}.
Their published scores therefore answer broader system-level questions than
our matched contrast. Differences in readers, graph construction, prompts,
context budgets, and metrics preclude a numerical ranking against our result.
Our evidence instead identifies what changes when the stored Memory text,
Memory vectors, query vectors, requested answer protocol, and evaluator remain
fixed.

The component results suggest that the complete difference is distributed
across the retrieval path. The semantic channel is the largest isolated
contributor. Entity fusion, sequence expansion, and speaker links add smaller
supported recall contributions. The raw-text comparison indicates that the
B-over-A recall advantage does not depend on the tested time annotation.
Sensitivity results also discourage presenting the primary parameters as
universally optimal. One lower-Entity-weight condition reduces recall, an
equal-weight condition is not distinguishable, and no tested sequence scale
survives family correction.

The configuration audit provides positive but bounded robustness evidence.
Across the four paper-eligible requested configurations, no requested-extractor
F1 or recall contrast rejects zero after Holm correction. This supports
limited empirical robustness across the tested GPT-3.5 and DeepSeek requested
extraction configurations, but it does not establish extractor equivalence,
independence from extraction, or arbitrary-model invariance. The embedding
result is mixed: F1 has no supported contrast, whereas recall falls for the
Doubao frozen artifacts relative to TES under the fixed, uncalibrated
0.30/0.70 fusion. Robustness therefore
applies to a specified outcome and tested configuration, not to unrestricted
model exchange without retuning or validation.

The category profile and cases are consistent with, but do not causally
identify, the intended retrieval mechanisms. In the audited single-hop and
temporal rows, Entity matching locates a relevant session and one-hop edges
add an adjacent turn. Open-domain rows can contain the annotated evidence
while their generated answers omit the reference rationale. Category 5 uses a
separate frozen option-generation and scoring branch, and the audited row
exhibits recall/F1 divergence under that branch. These observations place the
highest answer F1 in single-hop questions and higher recall than the original
Dialog@25 anchor in temporal and adversarial questions. They do not establish
a category-level causal effect or a protocol-matched cross-system advantage.

The cold/warm result exposes an operational tradeoff. Conversation Entity
extraction and Memory embedding are one-time costs, while uncached
question-Entity extraction dominates cold retrieval latency. Replaying the
same question set with complete caches requires no new provider request before
answer generation and reduces retrieval to milliseconds per question. This
measurement does not imply that unseen questions are free, or that another
machine, provider, or workload has the same latency.

\paragraph{Limitations and future work.}

First, formal evidence comes from ten conversations in one LoCoMo release.
Conversation-cluster bootstrap reflects uncertainty over these ten units but
does not establish cross-dataset generalization. Replication on additional
conversation-memory benchmarks is needed.

Second, no matched cutoff supports an overall final-answer F1 difference. The
contribution concerns retrieval coverage, not a general improvement in
generated answers. Alternative readers and evidence selectors require
separately controlled experiments.

Third, the official DRAGON reproduction gate did not pass. The local raw
DRAGON experiment changes semantic and query artifacts and remains diagnostic.
A future official-reference comparison requires parity with the pinned
upstream implementation.

Fourth, the requested answer-model name is recorded, but the provider did not
persist a separate actual-model identity in the formal snapshots. Category 5
also retains the frozen upstream unseeded option-order behavior. Future runs
should preserve provider-returned model identities when available and quantify
Category-5 generation variance separately. The robustness artifacts likewise
do not fully identify every provider deployment revision for extraction and
embedding. The matrix therefore supports outcome-specific empirical
robustness across the paper-eligible requested configurations represented by
their frozen artifacts, not robustness to fully identified provider
deployments; a strict deployment-level claim requires telemetry-complete
reruns.

Fifth, the two GPT-5 mini robustness cells have a frozen-versus-runtime
extraction-budget mismatch (2,500 versus 4,000 tokens), and 12 calls exceeded
the frozen limit. Their observed metrics are diagnostic only. A corrected
three-extractor robustness claim requires rebuilding and rerunning those cells
with one explicitly resolved budget; future graph-cache identities should also
bind extraction temperature and token budget rather than relying on source and
artifact hashes for retrospective audit.

Finally, system timing was measured on one machine and one provider endpoint.
The warm condition assumes complete reusable graph, index, query-vector, and
question-Entity artifacts for the same question set. Token usage is directly
measured, whereas the dollar conversion uses public prices accessed after the
experiment and is not a frozen metric. Prices, platform fees, and hardware
costs can change. Replication across hardware and providers is needed before
deployment-level latency claims.

\section{Conclusion}

Long-conversation question answering requires retrieval that preserves
relevant evidence without using test questions to build memory. Graph
retrieval may represent links that a flat index omits, but a fair test must
hold the dense representations, query vectors, answer generation, and
evaluation fixed. Our matched design addresses this evidence gap.

The matched comparison first finds that the Entity--Memory graph improves
official evidence recall over dense Memory retrieval on the evaluated LoCoMo
set, whereas the corresponding overall final-answer F1 difference is not
supported. Component contrasts then associate the recall difference with
semantic scoring, Entity fusion, chronological expansion, and speaker links.
Finally, the recall advantage persists across \topk{} 5--50 and the raw-text
input profile. Its magnitude is sensitive to the 0.10/0.90 Entity/Semantic
fusion setting, while no sequence-scale contrast survives family correction.
The paper-eligible requested-configuration matrix supports limited empirical
robustness of F1 and evidence recall across the tested GPT-3.5 and DeepSeek
extraction configurations. Embedding robustness is outcome-specific: neither
embedding F1 contrast rejects zero, whereas Doubao frozen artifacts reduce
recall by 2.02--2.29 points relative to TES. The corresponding intervals cross
the preregistered $-3$-point boundary, so the gate for material robustness
failure is not triggered, but embedding equivalence is not established.
Descriptive category ranges, supplemented by non-protocol external anchors,
place the highest answer F1 in single-hop questions, higher evidence recall
than Dialog@25 in temporal and adversarial questions, and the lowest answer F1
among Categories 1--4 in open-domain questions; adversarial raw F1 is lower
under its separate Category-5 branch. Cold/warm measurements also show that the
graph and retrieval artifacts can be reused for identical-question replay,
although unseen questions can require new provider calls.

This work is best viewed as a scoped retrieval study. The ten-conversation
sample, two paper-eligible requested extractors, two frozen embedding
artifacts, failed official DRAGON reproduction gate, single requested reader
alias/protocol, and single-environment latency measurement bound the claim.
Separating evidence coverage from answer quality provides a transferable
evaluation principle for future long-term conversational memory systems.

\section*{Reproducibility and Responsible Use}

\paragraph{Reproducibility.}
Every formal condition records its source commit, resolved configuration,
dataset and artifact SHA-256 values, isolated output path, query-cache usage,
provider-token telemetry, graph-input audit, and prompt budget. Query vectors
are complete and read-only during formal retrieval. The evaluation package is
vendored and hash-verified. Statistical reports and the final cost report have
independent byte-identical reproductions.

\paragraph{Responsible use.}
The graph uses conversation data, including speaker names and image captions,
which can contain personal information in real deployments. Systems applying
this method outside the benchmark should establish consent, retention,
deletion, and access-control policies. The present work evaluates benchmark
retrieval and does not study privacy attacks or sensitive-memory deletion.

\section*{Data and Code Availability}

LoCoMo data are available from the benchmark authors
\citep{maharana2024locomo}. Code, experiment manifests, validation records,
and manuscript evidence snapshots for this study are available at
\url{https://github.com/Sun668/em_graph_memory/tree/v1.0.8}. The exact manuscript snapshot
corresponds to the repository history recorded with the arXiv source bundle.

\begingroup
\small
\bibliographystyle{plainnat}
\bibliography{references}
\endgroup

\appendix
\onecolumn

\section{Resolved Primary Configuration}
\label{app:config}

\begin{table}[h]
\centering
\caption{Resolved settings for the primary B condition.}
\label{tab:config}
\begin{tabular}{@{}ll@{}}
\toprule
Dimension & Resolved value \\
\midrule
Dataset & LoCoMo ten-conversation release; 1,986 QA rows \\
Graph inputs & Session time, dialogue id, speaker, text, image caption \\
Excluded graph inputs & Questions, answers, evidence, categories, judges, predictions \\
Memory nodes & 5,882 \\
Conversation Entity extractor & \texttt{gpt-3.5-turbo}; temperature 0.3 \\
Memory embedding model & \texttt{text-embedding-3-small} \\
Answer model request & \texttt{gpt-3.5-turbo}; temperature 0; 32 output tokens \\
Primary context budget & \topk{} 25 \\
Entity / semantic weights & 0.30 / 0.70 \\
Entity relative-score threshold & 0.5 \\
Maximum matches per Entity key & 20 \\
Speaker-only dampening & 0.25 \\
Degree discount & Enabled \\
Sequence expansion scale & 0.5 \\
Query embeddings & Complete read-only artifact; L2-normalized float32 \\
Bootstrap & 10,000 paired QA and 10,000 conversation-cluster resamples \\
Bootstrap seed & 20260727 \\
\bottomrule
\end{tabular}
\end{table}

\section{Validity and Reproducibility Gates}

\begin{table}[h]
\centering
\caption{Checks applied before promotion of formal evidence.}
\label{tab:gates}
\begin{tabularx}{\textwidth}{@{}lXl@{}}
\toprule
Gate & Requirement & Outcome \\
\midrule
Graph construction & Conversation-only inputs; no QA-derived construction fields & Pass \\
Evaluator integrity & All 16 vendored files match the frozen SHA-256 manifest & Pass \\
Output isolation & One absent or empty directory per resolved formal condition & Pass \\
Query vectors & Complete ordered coverage; zero miss; zero live query embedding & Pass \\
Dense control & A and \(\bembed\) ordered contexts equal for all 1,986 rows & Pass \\
Prompt budget & Non-data scaffold no longer than 5,000 characters & Pass (2,410) \\
Statistical reproduction & Independent report build is byte-identical & Pass \\
Cost reproduction & Independent cost report build is byte-identical & Pass \\
Cache preservation & Quarantined artifacts retain frozen identities & Pass \\
\bottomrule
\end{tabularx}
\end{table}

\end{document}